\documentclass[10pt,twocolumn,letterpaper]{article}

\usepackage{iccv}
\usepackage{times}
\usepackage{epsfig}
\usepackage{graphicx}
\usepackage{amsmath}
\usepackage{amssymb}

\usepackage[breaklinks=true,bookmarks=false]{hyperref}

\iccvfinalcopy 

\ificcvfinal\fi

\begin{document}

\title{Roof material classification from aerial imagery}

\author{Roman Solovyev\\
Institute for Design Problems in Microelectronics of Russian Academy of Sciences\\
3, Sovetskaya Street, Moscow 124365, Russian Federation\\
{\tt\small turbo@ippm.ru}
}

\maketitle
\ificcvfinal\thispagestyle{empty}\fi

\begin{abstract}
This paper describes an algorithm for classification of roof materials using aerial photographs. Main advantages of the algorithm are proposed methods to improve prediction accuracy. Proposed methods includes: method of converting ImageNet weights of neural networks for using multi-channel images; special set of features of second level models that are used in addition to specific predictions of neural networks; special set of image augmentations that improve training accuracy. In addition, complete flow for solving this problem is proposed. The following content is available in open access: solution code, weight sets and architecture of the used neural networks. The proposed solution achieved second place in the competition "Open AI Caribbean Challenge".
\end{abstract}

\section{Introduction}

Some areas of the world are under significant risk of natural hazards such as earthquakes, hurricanes and floods; these acts of nature can have devastating consequences. One such area is the Caribbean. Disaster risk is especially great when houses and buildings do not meet modern construction standards, which is not uncommon in poor and informal settlements. Buildings can be retrofit to better prepare them for disaster, but the traditional method for identifying high-risk buildings involves visual inspection while going door to door by foot. This process can take many weeks if not months and cost millions of dollars.
To speed up this work, machine learning techniques can be helpful. So, the World Bank Global Program for Resilient Housing and WeRobotics teamed up to prepare aerial drone imagery of buildings across the Caribbean annotated with characteristics that matter to building inspectors \cite{worldbank2018}. A feature that is especially important is roof material. Actually, it is one of the main risk factors for earthquakes and hurricanes. 
To solve the problem of roof material classification, the competition – «Open AI Caribbean Challenge: Mapping Disaster Risk from Aerial Imagery» \cite{Drivendata2019} – was held at DrivenData site. The goal of this challenge was to create new rooftop classifiers using aerial imagery in St. Lucia, Guatemala, and Colombia. As known, machine learning models, in particular, deep convolutional neural networks, have now become the standard in image classification \cite{Rawat2017}, \cite{Sol2019}, \cite{Yu2017} as well as in the processing of satellite images and aerial photographs \cite{iglovikov2018ternausnetv2}, \cite{Sol2017}, \cite{buslaev2018fully}, \cite{seferbekov2018feature}, \cite{zhang2018automatic}. Machine learning models developed by the contestants can most accurately show the risk of disasters using drone images, and thus will help speed up building inspections and reduce their costs. Thanks to this, additional resources will be allocated for disaster preparation where the most disruptive hazards are expected.

\section{Problem statement}

The contestants received seven images of large areas in three countries: St. Lucia, Guatemala, and Colombia. One of these images is shown in Fig.1. 

For each image, polygons were provided for all building roofs, most of which were four-sided. For some of these rooftops, material was known; the others should be classified most accurately. 

In total, 5 possible roof materials were presented. They are listed in Table 1. Examples of images for each type of roof are shown in Fig. 2

\begin{figure}[ht!]
    \centering
    \includegraphics[width=0.8\linewidth]{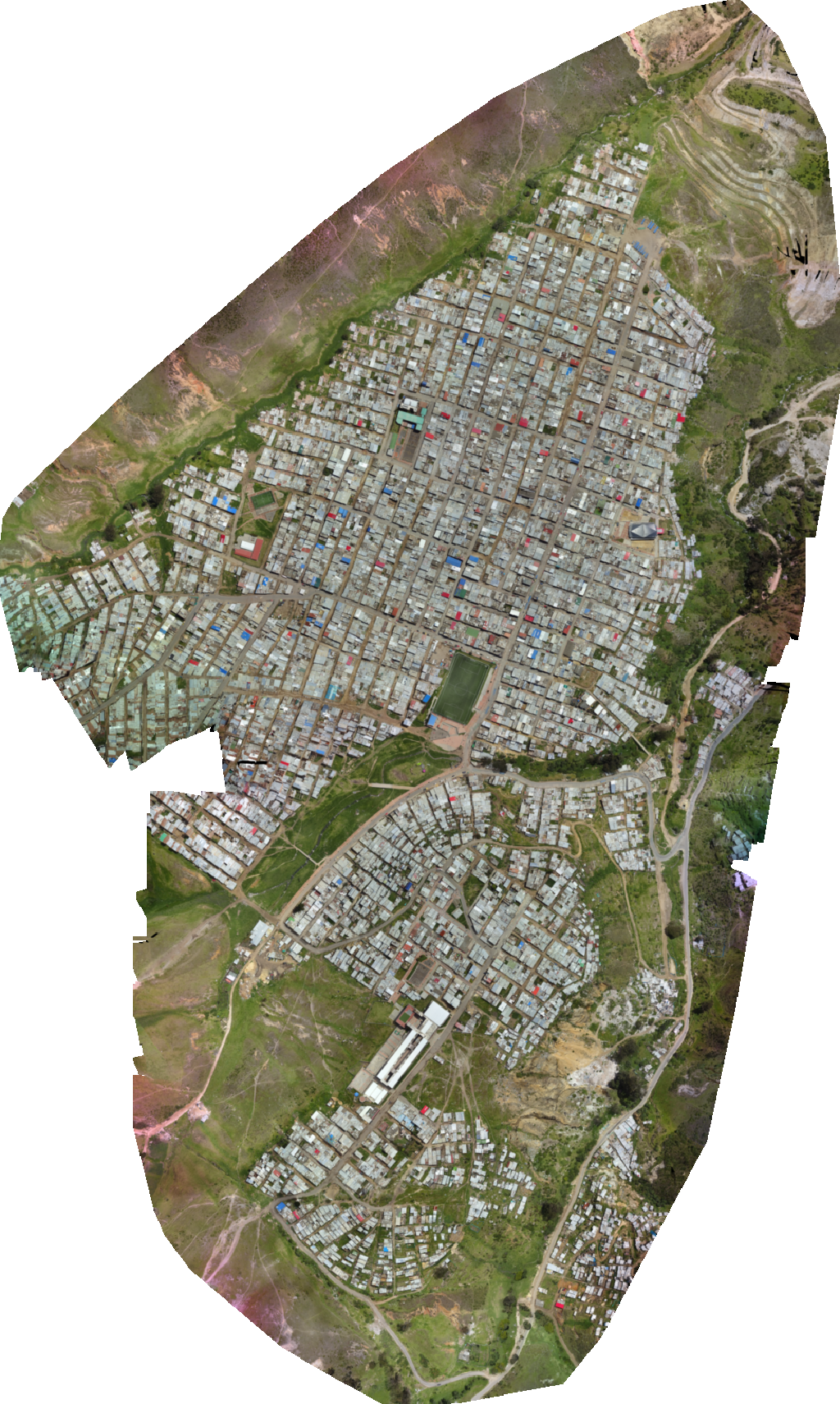}
    \caption{Photograph of Borde Rural, Colombia.}
    \label{fig:ctr-vis}
\end{figure}

\begin{table}[]
\begin{tabular}{|l|l|l|}
\hline
\multicolumn{1}{|c|}{\textbf{Roof material}} & \multicolumn{1}{c|}{\textbf{Description}}                                                                                                     & \multicolumn{1}{c|}{\textbf{\begin{tabular}[c]{@{}c@{}}Number of \\ buildings\end{tabular}}} \\ \hline
concrete\_cement                             & \begin{tabular}[c]{@{}l@{}}Roofs made of \\ concrete or cement\end{tabular}                                                                   & 1518                                                                                         \\ \hline
healthy\_metal                               & \begin{tabular}[c]{@{}l@{}}Roofs made of \\ corrugated metal, \\ galvanized metal \\ sheets or other \\ metallic materials\end{tabular}       & 14817                                                                                        \\ \hline
incomplete                                   & \begin{tabular}[c]{@{}l@{}}Roofs under construction, \\ damaged or partially \\ assembled roofs\end{tabular}                                  & 669                                                                                          \\ \hline
irregular\_metal                             & \begin{tabular}[c]{@{}l@{}}Metal roof with rust, \\ taps or damage. Such \\ roofs are at higher risk \\ during natural disasters\end{tabular} & 5241                                                                                         \\ \hline
other                                        & \begin{tabular}[c]{@{}l@{}}Roofs made of tiles or \\ other materials, and \\ painted roofs\end{tabular}                                       & 308                                                                                          \\ \hline
\end{tabular}
\caption{Roof types.}
\label{table:1}
\end{table}

Participants were to send a file with their solution, namely, the list of predicted probabilities that the roof belongs to a certain type for every house on each photograph and each roof type. To evaluate the solutions, log loss metric (1) was used: the lower the metric value, the better the solution. Metric formula is as follows:

\begin{equation}
    loss=-\frac{1}{N}\cdot\sum_{i=1}^N\sum_{j=1}^My_{ij}log(p_{ij}),
\end{equation}

where N is the total number of roofs in test examples, M is the number of classes (5, for this task), $\mathbf{y_{ij}}$ – 0 or 1 depending on whether the class corresponds to the given roof, $\mathbf{p_{ij}}$ – the predicted value of the probability that the given roof belongs to the given type.

For two of the seven images, only automatically generated non-validated mapping of roof material was available.

\begin{figure}[ht!]
    \centering
    \includegraphics[width=0.8\linewidth]{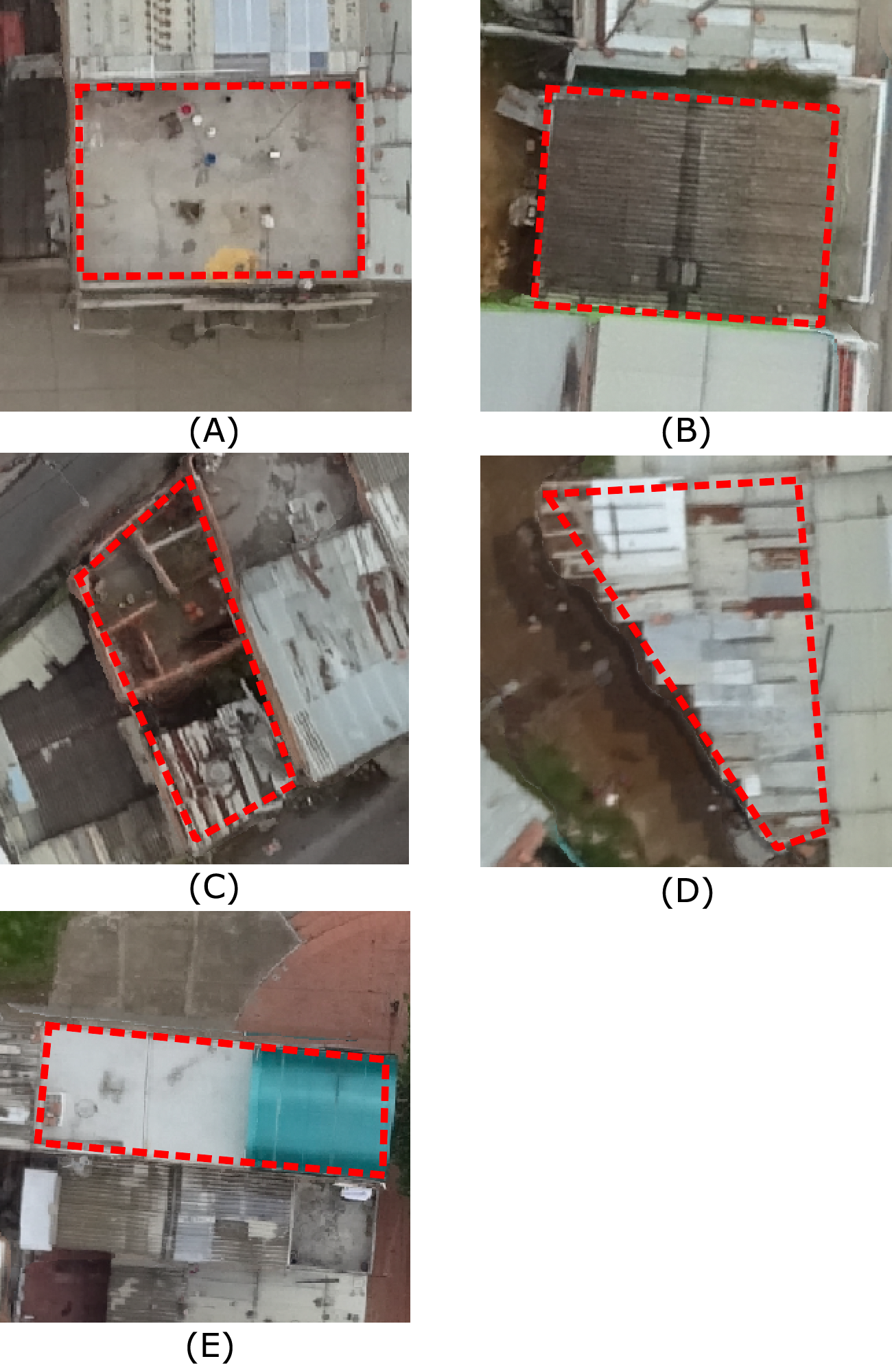}
    \caption{Roof types: \(A\) – concrete\_cement, \(B\) – healthy\_metal, \(C\) – incomplete, \(D\) – irregular\_metal, \(E\) – other.}
    \label{fig:ctr-vis}
\end{figure}

\section{Roofs classification for individual buildings}

To begin with, the images of each individual building were extracted together with corresponding polygonal masks; while doing this, some margin N from each edge of the mask was set (in the solution for this competition, N was set to 100) (see Fig. 3). It is clear that, for better classification the model should get roof masks, not only RGB images, as there can be more than one house and rooftop in the image.

\begin{figure}[ht!]
    \centering
    \includegraphics[width=0.8\linewidth]{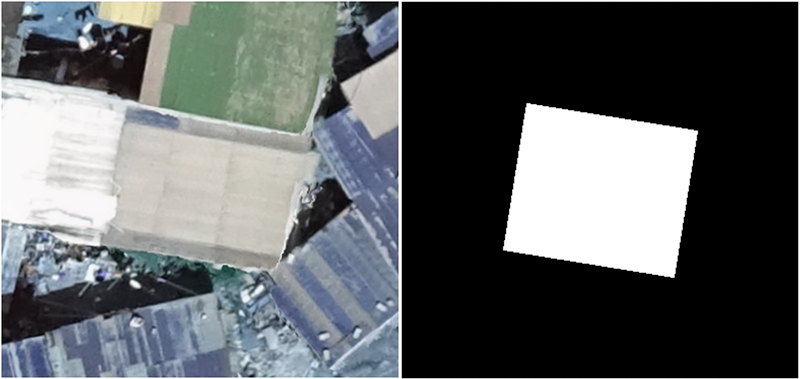}
    \caption{Roof image and mask.}
    \label{fig:ctr-vis}
\end{figure}

To solve classification problems, neural networks pre-trained on large sets of images are commonly used. Predominantly, these image sets are from the ImageNet database \cite{deng2009imagenet}. For all modern tools working with neural networks, there are neural network architectures for which ImageNet weights are available (ResNet \cite{he2016deep}, Densenet \cite{huang2017densely}, EfficientNet \cite{tan2019efficientnet}, MobileNet \cite{sandler2018mobilenetv2}, etc.). However, they all work with three-channel RGB images. To use four-channel images, we need some specialized method for adapting weights.

To solve this problem, it is enough to change parameters of the first convolutional layer. The convolutional layer parameters usually consist of "weight" and "bias" pairs. Bias is often not used in modern architectures. But even if it is used, the number of weighting factors for the bias depends on the number of output feature maps only. Therefore, we can keep the same bias when increasing the number of input channels. Now consider the convolutional layer weights.

The weights for Convolution2D layer are the four-dimensional matrix $\mathbf{W}$ with dimensions $\mathbf{K1\times K2\times M\times O}$ where $\mathbf{(K1, K2)}$ is the convolution kernel size, $\mathbf{M}$ is the number of input feature maps, $\mathbf{O}$ is the number of output feature maps.

Typical size of weight matrix for the first convolutional layer can be, for example, $\mathbf{W1 = (5\times5\times3\times32)}$. This record means that after applying convolution with $\mathbf{5\times5}$ kernel to a three-channel RGB image we get 32 feature maps at the layer output. Suppose that, for a neural network trained on ImageNet, we know this matrix $\mathbf{W1}$.

If a four-channel image arrives at the input of a neural network, the matrix of weights will have size: $\mathbf{W2 = (5\times5\times4\times32)}$. We have to construct $\mathbf{W2}$ from $\mathbf{W1}$ in such a way that neural network recognizes this four-channel input as similar to the three-channel one, and feature maps after this layer have similar distribution.

Several approaches are possible. The first variant is to form weights so that extra channels add zero contribution to feature maps. This can be done as follows:

	$\mathbf{M1 = 3, M2 = 4, W2[…] = 0}$
	
	$\mathbf{W2[:, :, :M1, :] = W1}$

The second variant is to recalculate the weights so that each of the channels contributes to feature maps obtained after the layer proportionally.

For each integer $\mathbf{j}$ from the interval from 0 to $\mathbf{M2}$, we perform the following operation:

	$\mathbf{W2[:, :, j, :] = M1 * w[:, :, j\%M1, :] / M2}$

For this problem, we used the second variant. 

This approach allows us to submit four-channel images to the model input and at the same time use high-quality ImageNet weights to further train the model on our data. This in turn increases the speed of learning and the quality of the resulting model.

\section{Splitting data into training and validation sets}

To begin with, we studied how the organizers splitted the data into training and test (validation) parts. Strictly speaking, the approach could be different. They could split the data by the maps, for example, appoint 5 maps for training and 2 for test, or by buildings – use some buildings within the same map for training and some for test. In the second case, the data could be grouped on area basis. The chosen way affects correct partitioning of data into training and validation sets. For this reason, locations of all training and test data were first plotted on a map and viewed visually (see Fig. 4).

\begin{figure}[ht!]
    \centering
    \includegraphics[width=0.8\linewidth]{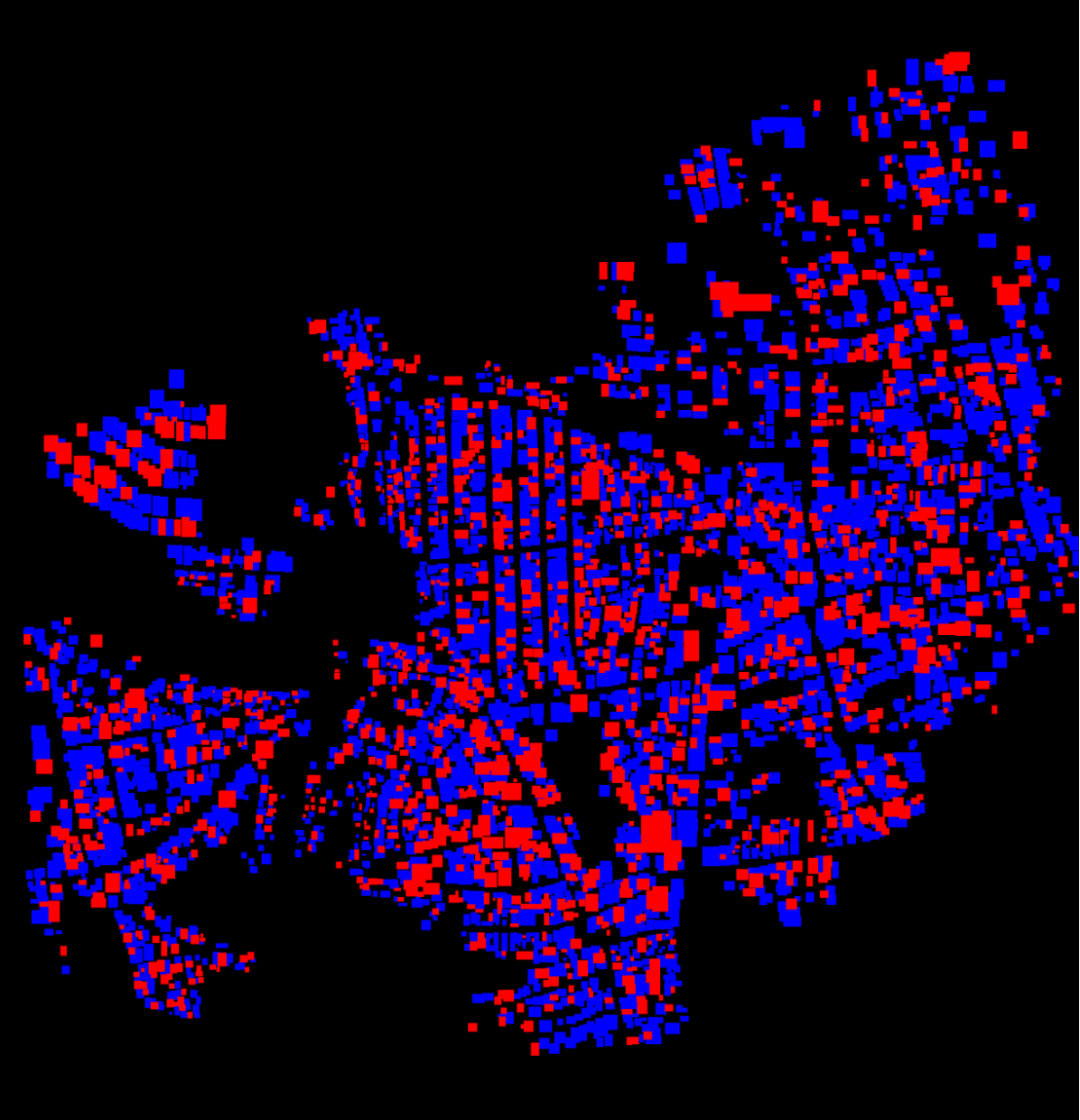}
    \caption{Roof distribution for training data (red) and test data (blue).}
    \label{fig:ctr-vis}
\end{figure}

From the obtained images, it became clear that the idea was to classify roofs that remained unmarked while partial visual inspection within a given map. We can also see that there were no explicit criteria for partitioning roofs into training and test sets within each single map. Therefore, assigning to training or validation set can be done randomly, separately for each image.

The organizers also stated that two pictures with automatic marking were not included into test markup. Therefore, they should not be used for validation. However, as part of the experiments, we found out that better use these images during training process.

\section{Global (Meta) features}

Unfortunately, neural networks for classification do not get the full data near the analyzed building. Therefore, it is possible (and necessary) to use second-level models for solving classification task. So, we use such feature as out of fold (OOF) predictions \cite{ML2019} from all neural networks in parallel with the other meta features for each individual building.

What can be used as useful attributes that can improve predictions? In this task, these were:

1) ID of the map

2) Roof type distribution for the nearest N buildings (neighbors), distance to each of the neighbors,  coordinates of the neighbors and the area of their roofs 

3) Roof type distribution for neighbors in the building vicinity for the given radius R

4) Location of the building in the map 

5) Roof area

The logic of feature choice is as follows:

1) In different regions, the popularity of roof materials is different. Some types of materials may occur more often. Within one map, there are richer areas where more expensive materials are used. This model may require coordinates of buildings. Also for large buildings some specific materials are used – this can be predicted by the roof area. 

2) Obviously, materials from neighboring houses are more likely to be similar. Therefore, statistics on neighbors can come in handy. 

3) People who made markings on different maps are prone to do the same mistake and choose some wrong option more often, and also classify objects as “other”. In this model, such meta-features as map number can help.

\section{Special methods of dataset augmentation}

Because the number of maps and buildings is small, neural networks get over-trained rather quickly. To increase solution quality and reduce possibility of over-training, a large set of augmentations was used. We used both conventional sets of augmentations and sets that we developed specially for this task.
We used such conventional augmentations as:

- Horizontal and vertical reflections, for which probabilities were selected so that all 8 variants meet equally often. 

- Image brightness shift for each of the three channels - RGB Shift. For this task, shift is applied randomly in the interval [-20; 20].

- Three types of random blur: Median Blur, Blur, GaussianBlur

- Random noise (including GaussNoise)

- A set of augmentations that shift image blocks in one or another way: Elastic Transform \cite{simard2003best}, Grid Distortion \cite{wigington2017data}, Optical Distortion \cite{fitzgibbon2001simultaneous}.

We also used the following special augmentations:

- As the images were cut out with a margin around the mask, images randomly cut from the initial image in the interval from 0 to 100 on each side were fed to the input of the neural network. After training, parts of the image with different margin of pixels from the mask (for example, 50) were cut out for images from validation and test sets. 

- In some cases, the mask provided by the organizers might not be very accurate. Therefore, a special augmentation was provided in which the mask was randomly shifted by several pixels and rotated by a small angle.

Examples of augmented images can be seen in Figures 5 and 6.

\begin{figure}[ht!]
    \centering
    \includegraphics[width=0.8\linewidth]{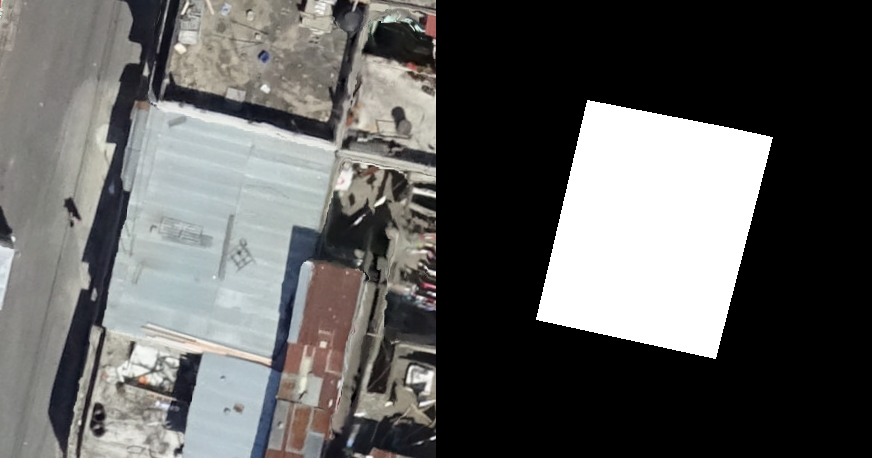}
    \caption{Original image and its mask.}
    \label{fig:ctr-vis}
\end{figure}

\begin{figure}[ht!]
    \centering
    \includegraphics[width=0.8\linewidth]{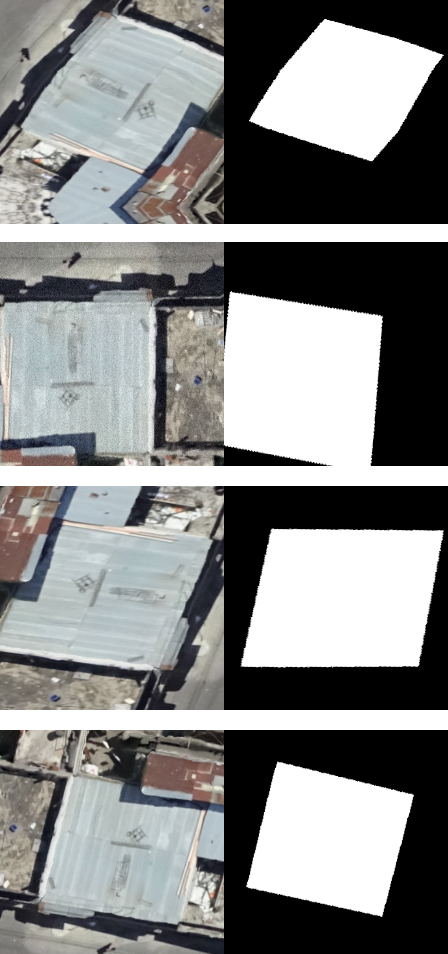}
    \caption{Random image augmentations.}
    \label{fig:ctr-vis}
\end{figure}

\section{Other specific features of training neural nets and second-level models}

1) To increase solution accuracy due to using ensemble, we prepared several different variants of neural networks with ImageNet weights. At the same time, in order to increase the variability of predictions, we significantly changed training options for each of these networks. These options included augmentation set, classification part structure, number of folds, and resolution of input images.
In total, 7 different variants of architectures were used:

- DenseNet121 and DenseNet169  \cite{huang2017densely} (input: 224x224)

- Inception Resnet v2 \cite{szegedy2017inception} (input: 299x299)

- EfficientNetB4 \cite{tan2019efficientnet} (input: 380х380)

- ResNet34, ResNet50 and SE-ResNext50 \cite{he2016deep}, \cite{hu2018squeeze} (input: 224x224) 

2) To prevent excessive over-training, we applied Dropout before the last layer of the networks. For different networks, values were selected in the range from 0 to 0.7. Also, in some cases additional Fully-Connected layer was used. 

3) At the inference (prediction) stage, the Test time augmentation (TTA) method was applied to test data, that is, each image was passed through the neural network several times. All 8 variants of reflections and rotations by 90 degrees were applied, and 4 different values of indentation from the boundaries were taken to cut out the center of the image (0, 25, 50, 75). That is, each image was run through the network 32 times. 

4) To train second level models, we used the packages XGBoost \cite{chen2016xgboost}, LightGBM \cite{ke2017lightgbm} and CatBoost \cite{prokhorenkova2018catboost}. Since the training set and the number of parameters were small, we preferred to run each package many times with random parameters rather than choose parameters carefully, the results were averaged.

5) An individual second level fully-connected neural network produced a bit worse result than GBM methods based on trees, and thus worsened the result of the ensemble.

General solution flow is shown in Fig. 7.

\begin{figure}[ht!]
    \centering
    \includegraphics[width=0.8\linewidth]{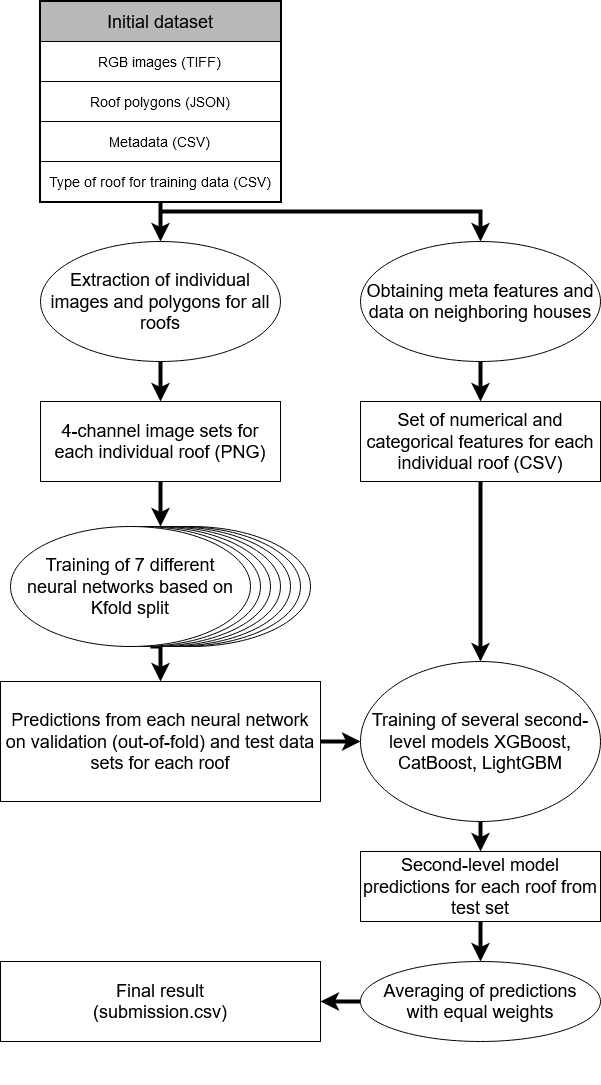}
    \caption{General pipeline of contest solution.}
    \label{fig:ctr-vis}
\end{figure}

\section{Results}

The solution described in this paper took 2nd place in the competition among more than 1,400 participants. The code and the set of weights for models are freely available on github \cite{Github2020}.

Table 2 shows the results of local validation and the results in the contest rating table separately for each neural network, for second-level models, as well as for the final result. Dashes in the table mean that the result is unknown (for example, such result was not sent to the contest server or the parameter was not calculated).

\begin{table}[]
\begin{tabular}{|l|c|c|}
\hline
\multicolumn{1}{|c|}{\textbf{Model}}                                                       & \textbf{\begin{tabular}[c]{@{}c@{}}Local \\ log\_loss / \\ accuracy\end{tabular}}                                           & \textbf{\begin{tabular}[c]{@{}c@{}}Public  \\ leaderboard\\ (log\_loss)\end{tabular}} \\ \hline
\begin{tabular}[c]{@{}l@{}}DenseNet121, \\ 224x224, \\ 5 Folds, TTA 32\end{tabular}        & 0.4303 / 0.8371                                                                                                             & 0.4082                                                                                \\ \hline
\begin{tabular}[c]{@{}l@{}}Inception Resnet \\ v2 299x299, \\ 5 Folds, TTA 32\end{tabular} & 0.4445 / 0.8317                                                                                                             & 0.4211                                                                                \\ \hline
\begin{tabular}[c]{@{}l@{}}EfficientNetB4 \\ 380x380, \\ 5 Folds, TTA 32\end{tabular}      & 0.4315 / 0.8379                                                                                                             & 0.4082                                                                                \\ \hline
\begin{tabular}[c]{@{}l@{}}DenseNet169, \\ 224x224, \\ 6 Folds, TTA 32\end{tabular}        & 0.4282 / 0.8373                                                                                                             & ---                                                                                   \\ \hline
\begin{tabular}[c]{@{}l@{}}ResNet34, \\ 224x224, \\ 5 Folds, TTA 32\end{tabular}           & 0.4543 / 0.8277                                                                                                             & ---                                                                                   \\ \hline
\begin{tabular}[c]{@{}l@{}}SE-ResNext50 \\ 224x224, \\ 5 folds, TTA 32\end{tabular}        & 0.4312 / 0.8375                                                                                                             & ---                                                                                   \\ \hline
\begin{tabular}[c]{@{}l@{}}Second-level \\ model CatBoost\end{tabular}                     & 0.3956 / ---                                                                                                                & 0.3700                                                                                \\ \hline
\begin{tabular}[c]{@{}l@{}}Second-level \\ model XGBoost\end{tabular}                      & 0.3702 / ---                                                                                                                & 0.3639                                                                                \\ \hline
\begin{tabular}[c]{@{}l@{}}Second-level \\ model LightGBM\end{tabular}                     & 0.3738 / ---                                                                                                                & ---                                                                                   \\ \hline
\begin{tabular}[c]{@{}l@{}}Second-level \\ model ensemble. \\ Final result\end{tabular}    & \begin{tabular}[c]{@{}c@{}}XGB \\ (Local: 0.3746) \\ + CatBoost \\ (Local: 0.3778) \\ + LGB \\ (Local: 0.3792)\end{tabular} & 0.3540                                                                                \\ \hline
\end{tabular}
\caption{Comparison of models by prediction loss.}
\label{table:2}
\end{table}

On completion, it turned out that training 7 neural networks was excessive – three networks were quite enough to get the best result. Second-level models significantly improved the result. Most likely, for practical use, one of the most accurate neural networks and an ensemble of second-level models will suffice. The use of the ensemble of second-level models is justified due to the very high performance of tree-based methods.

This work was supported by the Russian Science Foundation (RSF), grant No. 17-19-01645.

{\small
\bibliographystyle{ieee_fullname}
\bibliography{main}
}

\end{document}